%% file: paper.tex
\documentclass[10pt,twocolumn,letterpaper]{article}

\usepackage{iccv}
\usepackage{multirow}
\usepackage{times}
\usepackage{epsfig}
\usepackage{graphicx}
\usepackage{amsmath}
\usepackage{amssymb}
\usepackage{multirow}
\usepackage{amsmath}
\usepackage[b]{esvect}    

\usepackage{dsfont}
\DeclareMathOperator{\diag}{diag}  
\usepackage[title]{appendix}
\usepackage[accsupp]{axessibility}

\usepackage{algorithm}
\usepackage{algcompatible}

\usepackage[pagebackref=true,breaklinks=true,letterpaper=true,colorlinks,bookmarks=false]{hyperref}

\iccvfinalcopy

\ificcvfinal\fi

\def\mypar#1{\vspace{1mm}{\noindent\bf #1.}\hspace{1mm}}

\def\tab#1{Table~\ref{tab:#1}}
\def\sect#1{Section~\ref{sec:#1}}
\def\Eq#1{Eq.~(\ref{eq:#1})}
\def\app#1{Appendix~\ref{sec:#1}}

\begin{document}
\title{Generating Attribution Maps with Disentangled Masked Backpropagation}

\author{
Adria Ruiz \hspace{10mm} Antonio Agudo \hspace{10mm} Francesc Moreno-Noguer\\
Institut de Robotica i Informàtica Industrial, CSIC-UPC, Barcelona, Spain\\
{\tt\small \{aruiz,aagudo,fmoreno\}@iri.upc.edu}
}

\maketitle

\begin{abstract}
Attribution map visualization has arisen as one of the most effective techniques to understand the underlying inference process of Convolutional Neural Networks. In this task, the goal is to compute an score for each image pixel related to its contribution to the network output. In this paper, we introduce Disentangled Masked Backpropagation (DMBP), a novel gradient-based method that leverages on the piecewise linear nature of ReLU networks to decompose the model function into different linear mappings. This decomposition aims to disentangle the attribution maps into  positive, negative and nuisance factors by learning a set of variables masking the contribution of each filter during back-propagation. A thorough evaluation over standard architectures (ResNet50 and VGG16) and benchmark datasets (PASCAL VOC and ImageNet)  demonstrates that DMBP generates more visually interpretable attribution maps than previous approaches. Additionally, we quantitatively show that the maps produced by our method are more consistent with the true contribution of each pixel to the final network output.
\end{abstract}

\section{Introduction}
\input{introduction.tex}

\section{Related Work}
\label{sec:related}
\input{related.tex}

\section{Disentangled Masked Backpropagation}
\input{method.tex}

\input{method_bias_cnn}

\section{Experiments}
\input{experiments.tex}

\section{Conclusions}
We have presented Disentangled Masked Backpropagation, a novel gradient-based approach for attribution map generation. In contrast to previous methods, DMBP leverages on the piecewise linear nature of ReLU neural networks to disentangle positive, negative and nuisance factors from the attribution maps. Our experiments demonstrate that, compared to previous state-of-the-art methods, DMBP produces fine-grained attribution maps that are more visually interpretable and identify better the contribution of each pixel to the network output. Whereas we have focused on standard CNN architectures employing ReLU activations, our framework can also be applied to networks with other types of piecewise linear activations such as the Leaky-ReLU. Last but not least, other non-linearities such as the sigmoid or the hyperbolic tangent could be also introduced by modelling them with piecewise linear approximations.

\vspace{1mm}
\mypar{Acknowledgments} This work has been partially funded by the Spanish government with the project
MoHuCo PID2020-120049RB-I00. Adria Ruiz acknowledges financial support from MICINN (Spain) through the program Juan de la Cierva.

\clearpage

{\small
\bibliographystyle{ieee_fullname}
\bibliography{paper}
}

\clearpage
\begin{appendices}
\onecolumn
\input{supplementary.tex}

\end{appendices}

\end{document}

%% file: introduction.tex
Convolutional Neural Networks (CNNs) are ubiquitous in current state-of-the-art approaches for automatic visual understanding. 
Despite their outstanding performance in multiple tasks~\cite{garcia2017review,hossain2019comprehensive,zhao2019object}, they are still characterized as black-boxes whose internal inference rules  are difficult to interpret. As a consequence, the trustability of this type of models is limited and it holds back their broader adoption in applications such as autonomous driving~\cite{kim2017} or medical diagnosis~\cite{ching2018opportunities}, where it is crucial to ensure that model decisions are reliable and not based on data artifacts or biases.

\begin{figure}
\centering
\includegraphics[width=1.0\linewidth]{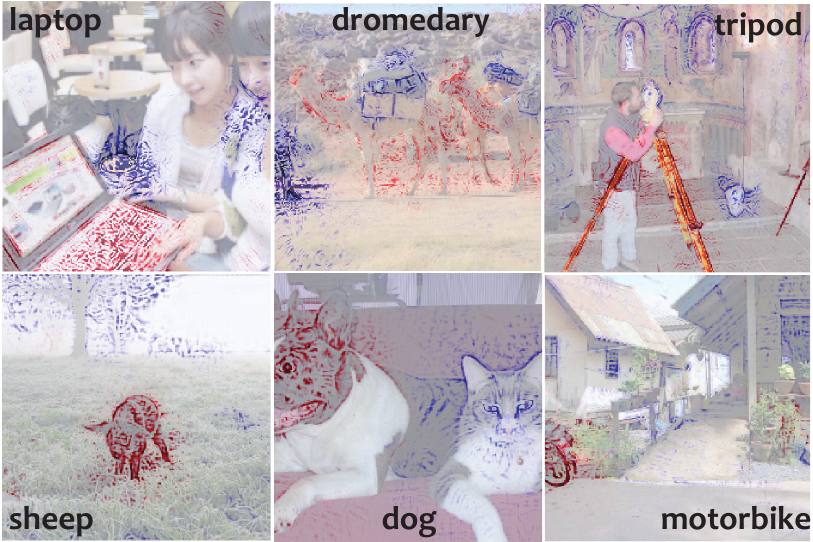}
\caption{{\bf Illustration of our proposed method for attribution map generation.} Given an input image, we want to obtain a score for each pixel estimating their contribution to the CNN output for a target label (in bold). Our approach is able to identify discriminative image pixels that contribute positively (red) or negatively (blue) to the prediction, and pixels corresponding to nuisance factors that have no effect on the output (white). For instance, in bottom-middle, positive attributions are assigned to the dog whereas negative scores correspond to pixels belonging to the cat. Additionally, no attributions are assigned to the non-discriminative background pixels. The disentanglement of this components produces fine-grained pixel-level attributions revealing the patterns used by the network during inference. We show that the generated attribution maps are more informative and visually interpretable than the ones obtained by previous methods. 
}
\label{fig:intro_fig}
\end{figure}

In this context, several strategies have been explored  to visualize the underlying rules guiding the model's decision process~\cite{zhang2018visual}. Attribution map generation is one of the most effective methods for this purpose~\cite{bach2015,petsiuk2018,selvaraju2019,springenberg2015,sundararajan2017,zeiler2014}. This task aims to assign an score to each individual input (\ie, pixels) determining their contribution to the final network output (\eg, the probability for a given class). By visualizing attribution maps, it is then easy to verify whether network inference is guided by intuitive rules such as the identification of discriminative image regions related to high-level semantic concepts (see Fig.~\ref{fig:intro_fig}). 

A promising approach to generate reliable attribution maps are gradient-based techniques~\cite{simonyan,sundararajan2017,zeiler2014,springenberg2015}. To determine the importance of each pixel, these methods use different mechanisms to  backpropagate the information from the output to the input image through the intermediate layers. An appealing property of gradient-based methods is that, compared to other approaches producing coarse and less informative attribution maps~\cite{petsiuk2018,selvaraju2019}, they can identify high-frequency patterns such as edges or textures. It has been shown that this information can be relevant to fully understand the network inference process~\cite{geirhos2018}.

In this paper, we introduce Disentangled Masked Backpropagation (DMBP). Similar to previous gradient-based approaches, our method uses backpropagation  to determine the contribution of each input pixel to the network output. However, DMBP addresses this task from a novel perspective. In particular, we use the fact that standard CNNs with ReLU non-linearities can be interpreted as piecewise linear functions where the input space is separated into different linear regions depending on the input \cite{xiong2020}. Using this observation, DMBP decomposes output's computation into different linear mappings that are used to disentangle nuisance, positive and negative factors from the attribution map. Whereas nuisance components refer to  information that have no effect on the network output, the latter  factors identify the discriminative pixels providing negative or positive evidences for the target label (see again Fig.~\ref{fig:intro_fig}). The different linear mappings are identified by decomposing the network gradient into different sub-components, which are identified by learning a set of variables masking network filters during backpropagation (see Fig.~\ref{fig:overview} for an overview). 

In our experiments, we validate the effectiveness of DMBP by providing qualitative and quantitative results over standard network architectures (ResNet50 and VGG16) and benchmark datasets (ImageNet and PASCAL VOC). The   results demonstrate that, compared to previous methods, attribution maps produced by DMBP  are more consistent with the true contribution of each pixel to the network output. Moreover, we show that our results are more informative and visually interpretable.

%% file: related.tex
\noindent\textbf{Attribution Map Generation} is one of the most effective strategies to understand the inference process of a CNN for a given input. For this purpose, perturbation-based methods measure the contribution of input pixels by observing the effect of excluding or including them during inference. These approaches use different mechanisms to generate binary masks defining image regions that are perturbed for network evaluation. Prediction Difference Analysis~\cite{zintgraf2017} and Occlusion~\cite{zeiler2014} use a sliding window approach to set to zero image patches and measure the effect on the CNN output. RISE~\cite{petsiuk2018} generates random binary masks and average them according to the target class probability. LIME~\cite{ribeiro2016model} and KernelShap~\cite{shap2017} weights image super-pixels according to a surrogate model that estimates the effect on the CNN output when they are removed. In contrast to these brute-force strategies, Meaningful~\cite{fonga17} and Extremal Perturbations~\cite{fong2019} pose this task as a learning problem where the mask is optimized to minimize the target label probability. However, while the previous methods have shown promising results, the generated attribution maps are sensitive to different hyper-parameters \cite{sam2020,ancona2017towards} controlling factors such as: (i) the type of image perturbation~\cite{fonga17}, (ii) the extracted super-pixels~\cite{shap2017,ribeiro2016model} (iii) the sampling process over the masks~\cite{zintgraf2017,petsiuk2018,ribeiro2016model} or (iiii) sparsity and smoothing constraints \cite{fong2019,fonga17}. These parameters can be difficult to validate in practice given the absence of an objective ground-truth.

Another popular approach to generate attribution maps is by exploiting the information contained in the intermediate network layers. In particular, Class Activation Maps~\cite{zhou2016learning} uses the weights of the final classifier to compute a linear combination of the feature maps in the last average pooling layer. GradCam~\cite{selvaraju2019} considered a similar approach with a linear combination  determined by the gradients of the output \wrt the last feature map. Score-CAM~\cite{wang2020} uses the intermediate layer activations to generate attribution maps following a similar strategy than perturbation-based methods. Full-gradient~\cite{srinivas2019} uses the gradient of the bias terms \wrt the output in order to generate attributions. More recently, Principal Feature Visualization~\cite{bakken2020} visualized the information of the last CNN layer through a PCA over the corresponding feature map. Finally, \cite{rebuffi20} combines multiple attribution maps generated from the gradient information  of the output \wrt the intermediate layer parameters. Despite these  approaches typically involve less hyper-parameters than perturbation-based methods, the information of intermediate layers is visualized by up-sampling it to the resolution of the original input image. As a consequence, the generated attribution maps are coarse-grained and do not reveal cues such as texture or edges that can be critical to understand the network inference process \cite{geirhos2018}. 

\begin{figure*}[ht]
\centering
\includegraphics[width=1.0\linewidth]{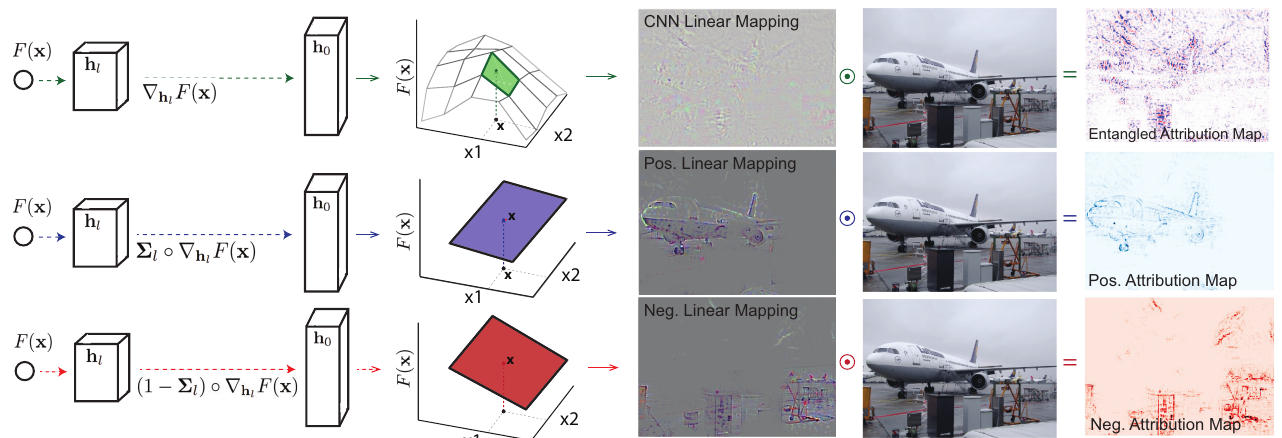}
\caption{{\bf Overview of Disentangled Masked Backpropagation for generating attribution maps.} {\bf Top:} Given a function $F(\mathbf{x})$ modelled by a CNN with ReLU non-linearities, the network output for a given image can be computed by applying a linear mapping over the input. This mapping is equivalent to the output gradients \wrt the input. Then, the attribution map indicating the contribution of each pixel can be computed as an element-wise multiplication of the mapping and the image pixels. As can be seen, however, this strategy typically produces noisy results that are difficult to visually interpret {\bf Middle and Bottom:} DMBP learns a set of variables weighting the contribution of each network filter during backpropagation. The optimization of these variables is guided by a loss which decomposes the original function into different linear mappings disentangling positive and negative attributions and removing nuisance factors from the attribution maps.
}
\label{fig:overview}
\end{figure*}

\noindent\textbf{Gradient based methods} for generating attribution maps are motivated by the fact that the gradient of a CNN output \wrt the input image is related with the contribution of each pixel to the final prediction. Based on this observation,~\cite{simonyan} proposed to directly use the network gradient to compute the importance of each pixel. However, the results of this approach tend to be too noisy to be easily interpreted. To overcome this limitation, several approaches average multiple gradients computed \wrt a set of modified input images. In particular, Integrated Gradients~\cite{sundararajan2017} considers a set of interpolations between the original and a reference input (\eg, a zero image). XRAI~\cite{kapishnikov19} applies this framework  to assign an score to different super-pixels. BIG~\cite{google2020} uses a set of blurred images as reference inputs. Finally, SmoothGrad~\cite{smilkov2017} averages multiple gradients resulting from evaluating different inputs corrupted with Gaussian noise. 

Rather than averaging multiple gradients, other approaches attempt to filter the non-relevant information during backpropagation by modifying the activation function derivatives~\cite{ancona2017towards}.
For instance, DeconvNet~\cite{zeiler2014} applies a ReLU activation to the gradient of each intermediate layer. Guided Backpropagation~\cite{springenberg2015} follows a similar strategy but the derivative is also zeroed if the input to the corresponding layer is negative. Guided GradCam~\cite{selvaraju2019} combines Guided Backpropagation attributions with the coarse-grained maps of GradCam. Finally, methods such as Excitation Backpropagation~\cite{zhang2016}, Layer-wise Relevance Propagation~\cite{bach2015}, DeepLift~\cite{shrikumar}, DeepShap \cite{shap2017}, Deep Taylor Decomposition~\cite{montavon2017explaining} or PatternAttribution~\cite{kindermans2017learning} employ different gradient computation rules to propagate attributions across layers. 

The proposed Disentangled Masked Backpropagation draws inspiration on the methods that use a gradient-based strategy to generate attribution maps. A fundamental difference, however, is that DMBP does not rely on hand-crafted backpropagation rules as in~\cite{zhang2016,bach2015,zeiler2014,springenberg2015}. Instead, it optimizes a set of variables masking the individual network filters during gradient computation. While learning function derivatives has been recently explored in LPR \cite{yang2019}, this method uses the modified gradients to generate a binary-mask similarly to perturbation-based methods. DMBP, instead, is the first approach to optimize backpropagation rules to explicitly decompose the network function into a set of linear mappings disentangling positive, negative and nuisance factors from the attribution maps. 

%% file: method.tex
In the following, we introduce the formal definition of attribution map used in our framework. Let us consider a linear model of the form $y=\mathbf{c}^T\mathbf{x}$, where $y \in \mathds{R}$ is the output (\eg, an score for a given class) and $\mathbf{c} \in \mathds{R}^{d_0}$ is a linear mapping applied to the input $\mathbf{x} \in \mathds{R}^{d_0}$ (\eg, a vectorized RGB image
).  Given any input-output pair $\{\mathbf{x},y\}$, we can compute an attribution map $\mathbf{a}\in \mathds{R}^{d_0}$ as $\mathbf{a}(\mathbf{x})=\mathbf{c} \odot \mathbf{x}$, where $\odot$ is the Hadamard product and $y=\sum_i \mathbf{a}_i$. As $\mathbf{x}$ is an image, $\mathbf{a}$ can be visualized in order to identify the contribution of each input pixel to the output. From now on, we use the previous definition of attribution map for DMBP and the rest of gradient-based approaches.

In this section, we show that standard ReLU networks model a linear function for each input $\mathbf{x}$ (\sect{linearizing}). For the sake of simplicity, we start considering networks with fully-connected layers and without bias terms. In \sect{decomposing}, we explain how DMBP uses this input-dependent linearization in order to disentangle positive, negative and nuisance factors from the attribution maps. Finally, in Sections \ref{sec:bias_incorporation} and \ref{sec:dmpb_cnn}, we generalize our framework to the case of networks with bias terms and CNNs,  respectively.

\subsection{Linearizing ReLU Neural Networks}
\label{sec:linearizing}

Let us consider a neural network with $L$ fully-connected layers defining a function $F: \mathds{R}^{d_0} \rightarrow \mathds{R}$ as:
\begin{equation}
    \label{eq:network_def}
    y =  F(\mathbf{x}) = \mathbf{w}^T \big[ f_{L}  \circ \ldots \circ f_{l} \circ  \ldots \circ f_{2} \circ f_1(\mathbf{x})\big],
\end{equation}
where $\mathbf{w} \in \mathds{R}^{d_L}$ represents a final filter computing an output from the last hidden layer (\eg, the target label score before applying a softmax). 
Additionally, each intermediate layer $f$ is defined as the composition of a linear function and a ReLU non-linearity as:
\begin{equation}
    \label{eq:mlp_hl}
    \mathbf{h}_l = f_l(\mathbf{h}_{l-1})=  \phi(\mathbf{W}_l \mathbf{h}_{l-1}),
\end{equation}
where $\mathbf{h}_l \in \mathds{R}^{d_l}$ are the intermediate layer activations, $\mathbf{W} \in \mathds{R}^{d_l \times d_{l-1}}$ and $\phi(\cdot)=\max(\cdot,0)$. 
Given previous definitions, we can express Eq.~\eqref{eq:mlp_hl} as:
\begin{equation}
\label{eq:mlp_linear_hl}
\mathbf{h}_l = \mathbf{\hat{W}}_l \mathbf{h}_{l-1} = \diag(\mathcal{H}(\mathbf{W}_l \mathbf{h}_{l-1})) \mathbf{W}_l \mathbf{h}_{l-1}, 
\end{equation}
where $\mathcal{H}(\cdot)$ denotes the Heaviside step function applied to all the elements in a vector. More intuitively, we model the ReLU operation as a diagonal binary matrix masking the subset of filters in $\mathbf{W}_{l}$ that produces negative elements via $\mathbf{W}_{l} \mathbf{h}_{l-1}$. As a result, we can express the composition of the linear mapping and the ReLU as a single matrix $\mathbf{\hat{W}}_l$.

From Eqs.~\eqref{eq:network_def} and~\eqref{eq:mlp_linear_hl}, it is easy to see that the network output is computed by applying a composition of linear transformations $\mathbf{\hat{W}}_l$ over the input $\mathbf{x}$ as:
\begin{equation}
\label{eq:mlp_out_linear}
y = \mathbf{w}^T \bigg[  \mathbf{\hat{W}}_{L} \dots \mathbf{\hat{W}}_{2} \mathbf{\hat{W}}_{1} \bigg] \mathbf{x} = \mathbf{w}^T \mathbf{H}_L \mathbf{x}.
\end{equation}

where $\mathbf{H}_L \in \mathds{R}^{d_L \times d_0}$. Note that for any linear function of the form $F(\mathbf{x}) = \mathbf{c}^T \mathbf{x}$, we have $\mathbf{c}=\nabla_\mathbf{x} F(\mathbf{x})$. 
Therefore, the vector $\mathbf{w}^T \mathbf{H}_L \in \mathds{R}^{d_0}$ in Eq.~\eqref{eq:mlp_out_linear} is equivalent to the gradient of the network's output \wrt the input. An attribution map $\mathbf{a}(\mathbf{x})\in \mathds{R}^{d_0}$ for a given image can thus be computed as:
\begin{equation}
    \label{eq:vanilla_map}
    \mathbf{a}(\mathbf{x}) = [\mathbf{w}^T \mathbf{H}_L] \odot  \mathbf{x}= \nabla_\mathbf{x} F(\mathbf{x}) \odot \mathbf{x}.
\end{equation}

\subsection{Attribution Map Disentanglement}
\label{sec:decomposing}
\mypar{Motivation} Using the output gradient to compute attribution maps was initially proposed in~\cite{simonyan}. However, this strategy typically produces noisy results that are difficult to visually interpret (see Fig.~\ref{fig:overview}-Top). To understand this phenomena, we need to analyse the role of the matrix $\mathbf{H}_L$ in \Eq{mlp_out_linear}. In particular, it can be interpreted as a linear mapping computing the  last layer features $\mathbf{h}_L$ from the input $\mathbf{x}$. However, note that the resulting features entangle both  discriminative and non-relevant information encoded by the network during inference. 
Nuisance components are thus also visualized in the attribution map, masking the discriminative factors that truly contribute to the model's output. 

Motivated by this observation, DMBP decomposes \Eq{mlp_out_linear} into three different terms:
\begin{equation}
    \label{eq:dmbp_decomposition}
    \mathbf{w}^T \mathbf{H}_L \mathbf{x}= 
    \mathbf{w}^T \big( \mathbf{H}^+_L+ 
    \mathbf{H}^-_L+
    \mathbf{H}^\sim_L \big) \mathbf{x},
\end{equation}
where ${\mathbf{H}}^+_L$ and ${\mathbf{H}}^-_L$ are linear mappings producing features that contribute positively and negatively to the output, respectively. In contrast, $\mathbf{H}^\sim_L$ aims to extract the non-discriminative features. 

Using this decomposition, an attribution map without nuisance factors can be computed as:
\begin{equation}
    \label{eq:decomposed_map}
    \mathbf{a}(\mathbf{x}) = [\mathbf{w}^T \mathbf{H}^+_L] \odot  \mathbf{x} + [\mathbf{w}^T \mathbf{H}^-_L] \odot  \mathbf{x}.
\end{equation}

\mypar{Filter decomposition} To obtain the decomposition in \Eq{dmbp_decomposition},  we use the fact that the feature extractor $\mathbf{H}_L$ in \Eq{mlp_out_linear} is defined as a product of matrices $\mathbf{\hat{W}}_l$. We can therefore decompose each of these linear mappings as:
\begin{equation}
    \mathbf{\hat{W}}_{l} = \mathbf{\hat{W}}^+_{l} + \mathbf{\hat{W}}^-_{l} = \mathbf{\Sigma}_l \mathbf{\hat{W}}_{l} + (\mathbf{I}-\mathbf{\Sigma}_l) \mathbf{\hat{W}}_{l},
\end{equation}
where $\mathbf{I}$ is the identity and $\mathbf{\Sigma}_l=\diag(\boldsymbol \sigma_l)\in \mathds{R}^{d_l \times d_{l}} $ is a diagonal matrix whose  entries $\boldsymbol \sigma_l \in [0,1]^{d_l}$ are vectors of learnable parameters.

Denoting $\boldsymbol \sigma = \{ \boldsymbol \sigma_L,\dots,\boldsymbol \sigma_1 \}$, the network output in \Eq{mlp_out_linear} can be explicitly decomposed into positive, negative and nuisance terms as:
\begin{align}
\label{eq:dmbp_decomposition2}
y &= \mathbf{w}^T \big( \mathbf{H}^+_L+
    \mathbf{H}^-_L+
    \mathbf{H}^\sim_L \big)\mathbf{x} \nonumber  \\ &= {y^+}(\boldsymbol \sigma) + {y^-}(\boldsymbol \sigma)  + {y^\sim}(\boldsymbol \sigma) \nonumber \\
&=  \mathbf{w}^T \bigg[  \mathbf{\Sigma}_L \mathbf{\hat{W}}_{L} \dots \mathbf{\Sigma}_1 \mathbf{\hat{W}}_{1} \bigg] \mathbf{x}  \nonumber \\
& \hspace{3mm} + \mathbf{w}^T \bigg[  (\mathbf{I}-\mathbf{\Sigma}_L) \mathbf{\hat{W}}_{L} \dots (\mathbf{I}-\mathbf{\Sigma}_1) \mathbf{\hat{W}}_{1} \bigg] \mathbf{x} \nonumber \\
& \hspace{3mm} + \mathbf{w}^T \mathbf{H}^\sim_{L} \mathbf{x},
\end{align}
where the masks $\mathbf{\Sigma}_l$ and $(\mathbf{I}-\mathbf{\Sigma}_l)$ select the set of filters for each layer producing features that have a positive or negative effect on the output, respectively. In contrast, $\mathbf{H}^\sim_{L}$ models the non-discriminative features.

\mypar{Learning objective} In order to learn the optimal parameters $\boldsymbol \sigma$ for a given input image $\mathbf{x}$, DMBP optimizes: 
\begin{equation}
\label{eq:loss}
\min_{\boldsymbol \sigma_1,\boldsymbol \sigma_2,\dots,\boldsymbol \sigma_L} y^-(\boldsymbol \sigma) - y^+(\boldsymbol \sigma) + \lVert y^\sim(\boldsymbol \sigma)\rVert_1,
\end{equation}
where we aim to maximize and minimize the positive and negative terms in \Eq{dmbp_decomposition2}, respectively. Additionally, the term $\lVert y^\sim(\boldsymbol \sigma)\rVert_1$ encourages  nuisance factors to have a negligible effect on $y$. During optimization, we ensure the constraint $\boldsymbol \sigma_l \in [0,1]^{d_l}$ by applying a sigmoid function over the set of learned scalar parameters.

\mypar{Optimization via masked backpropagation} 
Analogously to \Eq{mlp_out_linear}, the positive and negative terms  in \Eq{loss} are linear \wrt the input and thus, they can be expressed as:
\begin{equation}
\label{eq:pos_term}
y^+(\boldsymbol \sigma)=\nabla_\mathbf{x} F^+(\mathbf{x})^T \mathbf{x} \hspace{2.5mm} , \hspace{2.5mm} y^-(\boldsymbol \sigma)=\nabla_\mathbf{x} F^-(\mathbf{x})^T \mathbf{x}
\end{equation}
where $\nabla_\mathbf{x}  F^+(\mathbf{x})$
is obtained by performing a backward pass over the network while multiplying the filters of each layer by $\mathbf{\Sigma}_l$. Similarly, $\nabla_\mathbf{x}  F^-(\mathbf{x})$ can be obtained with another backward pass using $(\mathbf{I}-\mathbf{\Sigma}_l)$. Finally, the nuisance term does not require any explicit computation since it can be estimated through the network output and the previous computed terms as 
$y^\sim(\boldsymbol \sigma) = y - y^+(\boldsymbol \sigma) - y^-(\boldsymbol \sigma)$. Upon the definition of these computations,  the parameters $\boldsymbol \sigma_l$ for each layer can be optimized by minimizing the loss function in \Eq{loss} using standard gradient descent.


%% file: method_bias_cnn.tex
\subsection{Incorporating Bias Terms}
\label{sec:bias_incorporation}
In previous sections, we have obviated the biases terms for each filter that are typically used in neural networks. To take them into account, we shall modify \Eq{mlp_linear_hl} as:
\begin{equation}
    \label{eq:mlp_bias_hl}
    \mathbf{h}_l = f_l(\mathbf{h_{l-1}}) =  \phi(\mathbf{W}_l \mathbf{h}_{l-1}  + \mathbf{b}_{l}),
\end{equation}
where $\mathbf{b}_{l}$ is the bias term of the filter $\mathbf{W}_{l}$. 

Similar to \Eq{mlp_out_linear}, it is easy to show that, in this case, the network function can be linearized for a given input as: 
\begin{equation}
    \label{eq:bias_terms}
    y =  \mathbf{\hat{w}}^T \bigg[  \mathbf{\hat{W}}_{L} \dots \mathbf{W}_{1} \bigg] \mathbf{x} + \sum_{l=2}^L \mathbf{\hat{w}}^T \bigg[  \mathbf{\hat{W}}_{L} \dots \mathbf{\hat{W}}_{l} \bigg] \mathbf{b}_{l-1} 
\end{equation}
where $\mathbf{\hat{W}}_{l} = \mathbf{W}_{l} \diag(\mathcal{H}(\mathbf{h}_{l-1}  + \mathbf{b}_{l-1}))$, and $\mathbf{\hat{w}} = \mathbf{\hat{w}}  \diag(\mathcal{H}(\mathbf{h}_{L}))$. See \app{sup_bias_details} for more details. Note that the output $y$ is now obtained by applying a set of linear mappings over the input $\mathbf{x}$ and each bias term $\mathbf{b}_l$. Yet, the resulting function is again linear with respect to $\mathbf{x}$ and $\mathbf{b}_{0:L}$. Therefore, we can also express \Eq{bias_terms}  using the output gradients \wrt the input and biases as:
\begin{equation}
\label{eq:bias_grad}
y = \nabla_\mathbf{x}F(\mathbf{x})^T \mathbf{x} + \sum_{l=1}^L \nabla_\mathbf{b_l} F(\mathbf{x})^T \mathbf{b}_l.
\end{equation}
While the previous expression was firstly developed in \cite{srinivas2019} by using a different derivation, we use \Eq{bias_grad} in order to compute the DMBP decomposition in \Eq{dmbp_decomposition2} for neural networks with bias terms. Concretely,  we follow the same procedure described in Sec.~\ref{sec:decomposing}. However, the gradients \wrt the biases need to be computed during the two independent backward passes using $\mathbf{\Sigma}$ and ($\mathbf{I} -\mathbf{\Sigma})$. This is required to compute the contribution of the bias terms in \Eq{bias_grad}. 

\subsection{DMBP for Convolutional Neural Networks}
\label{sec:dmpb_cnn}
\mypar{Masking Convolutional Filters} CNNs compute   intermediate feature maps by applying a convolutional layer of the form $\mathbf{h_l}=\phi(\mathbf{W}_l* \mathbf{h}_{l-1})$. In this case, the composition of the convolutional operator and the ReLU can be also expressed as a single linear mapping:
\begin{equation}
    \label{eq:conv_operator}
   \mathbf{h_l}=\phi(\mathbf{W}_l * \mathbf{h}_{l-1}) = \mathcal{H}(\mathbf{W}_l * \mathbf{h}_{l-1}) \odot (\mathbf{W}_l * \mathbf{h}_{l-1}).
\end{equation}

The DMBP decomposition in \Eq{dmbp_decomposition2} can be also applied to convolutional layers as follows. Firstly, the positive term ${y}^+(\boldsymbol \sigma)$ can be obtained by multiplying $\mathbf{\Sigma}_l$ by the resulting feature maps after each convolutional and ReLU:
\begin{equation}
    \label{eq:cnn_decomp}
    \mathbf{h_l}= \mathbf{\Sigma}_l \circ \mathcal{H}(\mathbf{W}_l * \mathbf{h}_{l-1}) \odot (\mathbf{W}_l * \mathbf{h}_{l-1}),
\end{equation}
where $\mathbf{\Sigma}_l$ is a tensor of the same dimension than $\mathbf{h}_l$. Intuitively, this is equivalent to mask the applied filters $\mathbf{W}_l$ independently for each spatial position and channel of the input feature map. In this manner, the term ${y}^+(\boldsymbol \sigma)$ can be also computed as in \Eq{pos_term}, where $\nabla_\mathbf{x} F^+(\mathbf{x})$ is obtained with a single backward pass over the network by modifying the gradients for each intermediate layer as $\mathbf{\Sigma}_l \circ \nabla_{\mathbf{h}_l} F(\mathbf{x})$. Similarly, the negative term ${y}^-(\boldsymbol \sigma)$ can be computed using $(\mathbf{I}-\mathbf{\Sigma}_l)$ during backpropagation. Pseudo-code for DMBP optimization is provided in \app{sup_pseudocode}.

\mypar{Applying DMBP for other layers} Besides convolutions and ReLUs, standard CNNs also incorporate Batch Normalization (BN)~\cite{ioffe2015batch} or residual layers~\cite{he2016deep}. Fortunately,  the use of  these layers does not requires any modification into our proposed framework. The reason is that they can also be modelled as linear mappings over the input and thus, the network function can still be linearized as in \Eq{bias_grad}. During evaluation, BN can be fused with its previous convolution by modifying its filters and bias terms.\footnote{https://nenadmarkus.com/p/fusing-batchnorm-and-conv/}. On the other hand, residual layers of the form $\mathbf{h_l}=\phi(\mathbf{W}_l \mathbf{h}_{l-1}) + \mathbf{h}_{l-1}$ can  be represented by a linear mapping $\mathbf{h_l}=(\mathbf{\hat{W}}_l + \mathbf{I}) \mathbf{h}_{l-1}$, where $\mathbf{\hat{W}}_l$ is defined in \Eq{mlp_linear_hl}.

%% file: experiments.tex
\begin{figure*}[t]
\centering
\includegraphics[width=1.0\linewidth]{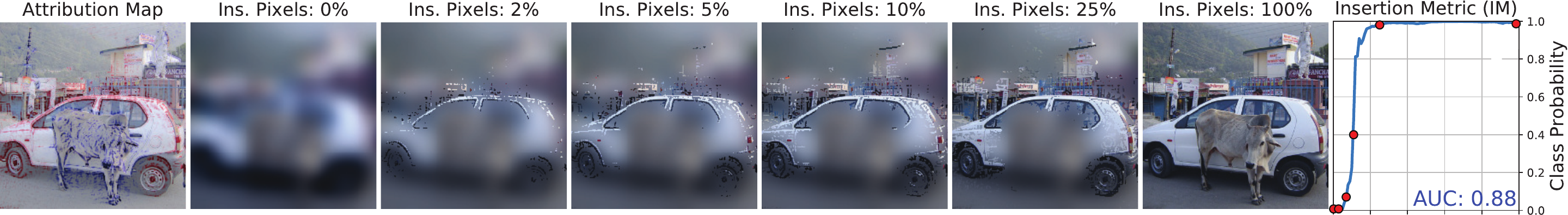}
\caption{{\bf Illustration of the Insertion Metric (IM) used in our experiments.} See text for details.
}
\label{fig:insertion_metric}
\end{figure*}

\mypar{Datasets and models} We conduct our experiments over benchmark datasets and architectures for image classification. In particular, we use the validation sets of ImageNet~\cite{russakovsky2015imagenet} and VOC2012~\cite{everingham2015pascal}. 
As baseline models, we use two extensively used CNN architectures: ResNet50~\cite{he2016deep} and VGG16~\cite{simonyan2014very}. Over ImageNet, we use the pretrained models in the \href{https://pytorch.org/docs/stable/torchvision/}{Torchvision} library for both architectures. In VOC2012, we use the models trained in~\cite{zhang2016}. 

\mypar{Baselines} We compare DMBP with $11$ previous approaches for attribution map generation, including state-of-the-art methods. Given that DMBP is a gradient-based approach, we focus on the comparisons with previous methods following this strategy: Grad~\cite{simonyan}, Integrated Gradients (IG)~\cite{sundararajan2017}, Smooth Gradients (SG)~\cite{smilkov2017}, Blurred Integrated Gradients (BIG)~\cite{google2020}, DeepLift (DL)~\cite{shrikumar2017}, Gradient Backpropagation (GBp)~\cite{springenberg2015} and Guided GradCam (GGC)~\cite{selvaraju2019}. As a reference, we also compare our method with GradCam (GC)~\cite{selvaraju2019} and FullGradients (FG)~\cite{srinivas2019}, which use the information in intermediate layers to compute the attribution maps. Finally, we also provide comparisons with the perturbation-based approaches RISE~\cite{petsiuk2018} and LPR~\cite{yang2019}.

\mypar{Implementation details and hyper-parameters} We use a \href{https://pytorch.org/}{PyTorch} implementation for DMBP and the rest of compared methods. For {BIG} \footnote{https://github.com/PAIR-code/saliency} and {FG} \footnote{https://github.com/idiap/fullgrad-saliency}, we integrate the code provided by the authors. We use our own implementation of LPR given that no code is publicly available. For the rest of methods, we use the implementations in the \href{https://captum.ai/}{Captum.ai} and \href{https://facebookresearch.github.io/TorchRay/}{TorchRay} libraries. The hyper-parameters for all the compared methods are set to the default values suggested in the original papers. For DMBP, we use RMSProp with a learning rate of $0.01$ as an optimizer to minimize the loss in \Eq{loss}. No weight decay is applied. A total number of $200$ iterations are executed during gradient-descent. The optimization for a given $224 \times 224$ input image requires $\sim 20s$ on a NVIDIA 2080 Ti GPU. This time is higher than the needed by other gradient-based approaches. However, our primary goal is to generate accurate attribution maps rather than their efficient computation. Our code implementing DMBP is publicly available in this  \href{https://gitlab.com/adriaruizo/dmbp_iccv21}{repository}.

\input{Tables/ablation_table}

\subsection{Evaluation Metrics}
Evaluating attribution maps is challenging given the absence of objective ground-truth. Previous works have attempted to evaluate them by using object bounding-boxes provided by human annotators~\cite{selvaraju2019,zhang2016} or conducting user-studies~\cite{shap2017}. However, human-based evaluation can be flawed and misleading~\cite{adebayo2018sanity}, given that the perceived discriminative pixels can differ between humans and CNNs.

\input{Tables/soa_table2}

\mypar{Insertion Metric (IM)}  To overcome this limitation, we use the insertion metric proposed in~\cite{petsiuk2018}.  This metric does not rely on human annotations and computes a score for each image as follows. Given an attribution map generated for a target label, the pixels are ordered in decreasing order according to their attribution value. Then, an increasing percentage of pixels is added iteratively to a reference blurred image and the probability of the target label is evaluated with the network. Finally, the AUC over the probabilities is computed to obtain a single score for the image (see Fig.~\ref{fig:insertion_metric} for an illustration). Intuitively, this metric attempts to measure if the pixels with large attribution values contribute positively to the network output. Different from \cite{petsiuk2018} where the attribution maps were generated for the label with a highest probability, we use the ground-truth classes provided in each dataset as the target label. This results in a more challenging problem because the class can be predicted with a low probability. Additionally, using the ground-truth annotations as target labels allows us to use the following complementary metric to IM.

\mypar{Complementary Insertion Metric (cIM)} In IM, the first inserted pixels are the ones with highest attributions. As a consequence, this metric is not appropriate to evaluate the information provided by pixels with negative scores. We expect such pixels to correspond to regions that are discriminative but provide negative evidences for the target label. To evaluate these attributions, we use an alternative metric that can be applied to images annotated with multiple classes and which we refer to as Complementary Insertion Metric (cIM). In particular, we follow the same procedure as in IM but the pixels with lower attributions are inserted first. Then, the AUC is computed by evaluating the probabilities for all the ground-truth labels that are different from the target class used to compute the attribution map. Therefore, a high cIM indicates that pixels with negative attributions correspond to discriminative regions providing positive evidences for the complementary classes in the image. Consequently, these regions provide negative evidences for the target class.  We do not compute cIM for ImageNet given that images are only labelled with a single class.

\subsection{Evaluating Attribution Map Disentanglement}
In this experiment, we evaluate the effect of disentangling positive, negative and nuisance factors on the generated attribution maps. For this purpose, we compare different variants of DMBP optimized with ablated versions of our loss defined in \Eq{loss}. Concretely, we use: (i) A loss maximizing only the positive term ${y}^+$. (ii) The same objective but also minimizing ${y}^-$. (iii) The original loss that also takes into account the nuisance term ${y}^\sim$. From now on, we refer to these approaches as DMPB$^+$, DMBP$^{+,-}$ and DMBP$^{\text{All}}$, respectively. Note that DMPB$^+$ is optimized to identify only positive factors; DMBP$^{+,-}$ aims to disentangle positive and negative factors;  DMBP$^{\text{All}}$ also seeks to remove nuisance factors. In addition, we also evaluate attribution maps generated by the vanilla approach in \Eq{vanilla_map}, where factors are not disentangled. We refer to this method as ND. For a faster experimentation, we use a subset of $5$K images for ImageNet with five random images per class.

\begin{figure}[t]
\centering
\includegraphics[width=1.0\linewidth]{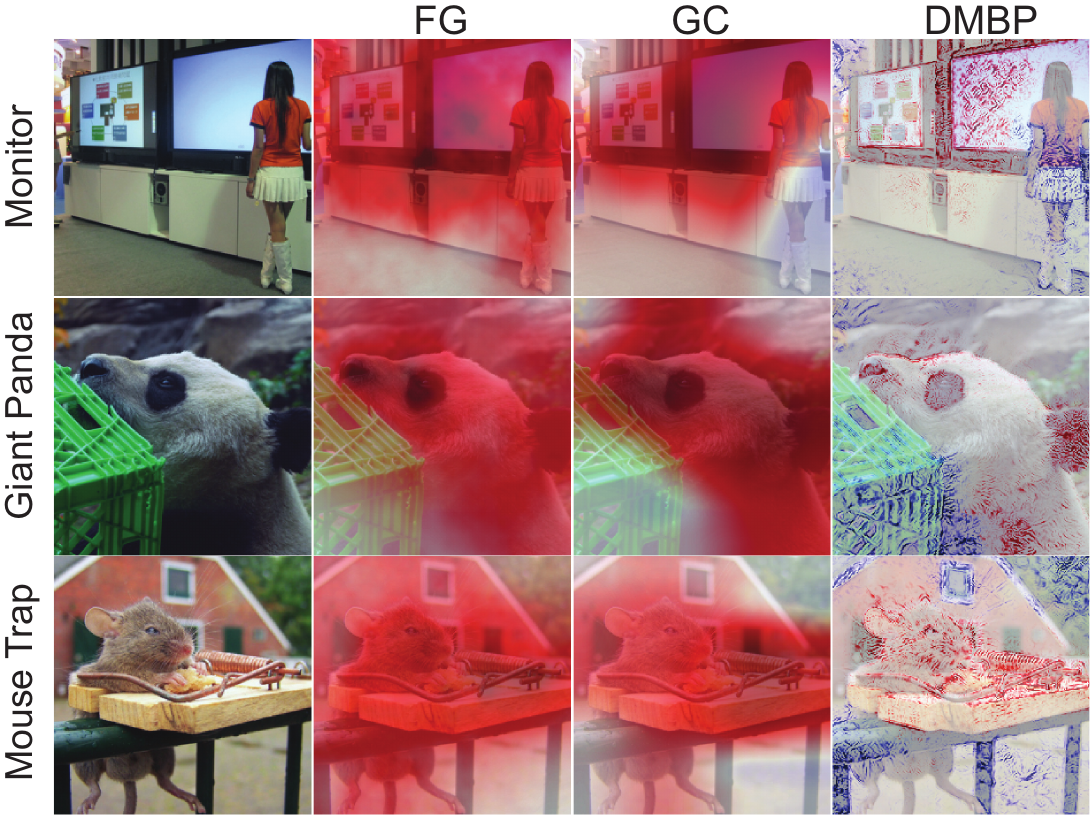}
\caption{{\bf Comparison between attribution maps generated by FG, GradCam and the proposed DMBP.} The latter generates fine-grained pixel attributions that are more informative and interpretable than the coarse results produced by the other methods.
}
\label{fig:coarse_com}
\end{figure}

\begin{figure*}[ht]
\centering
\includegraphics[width=0.99\linewidth]{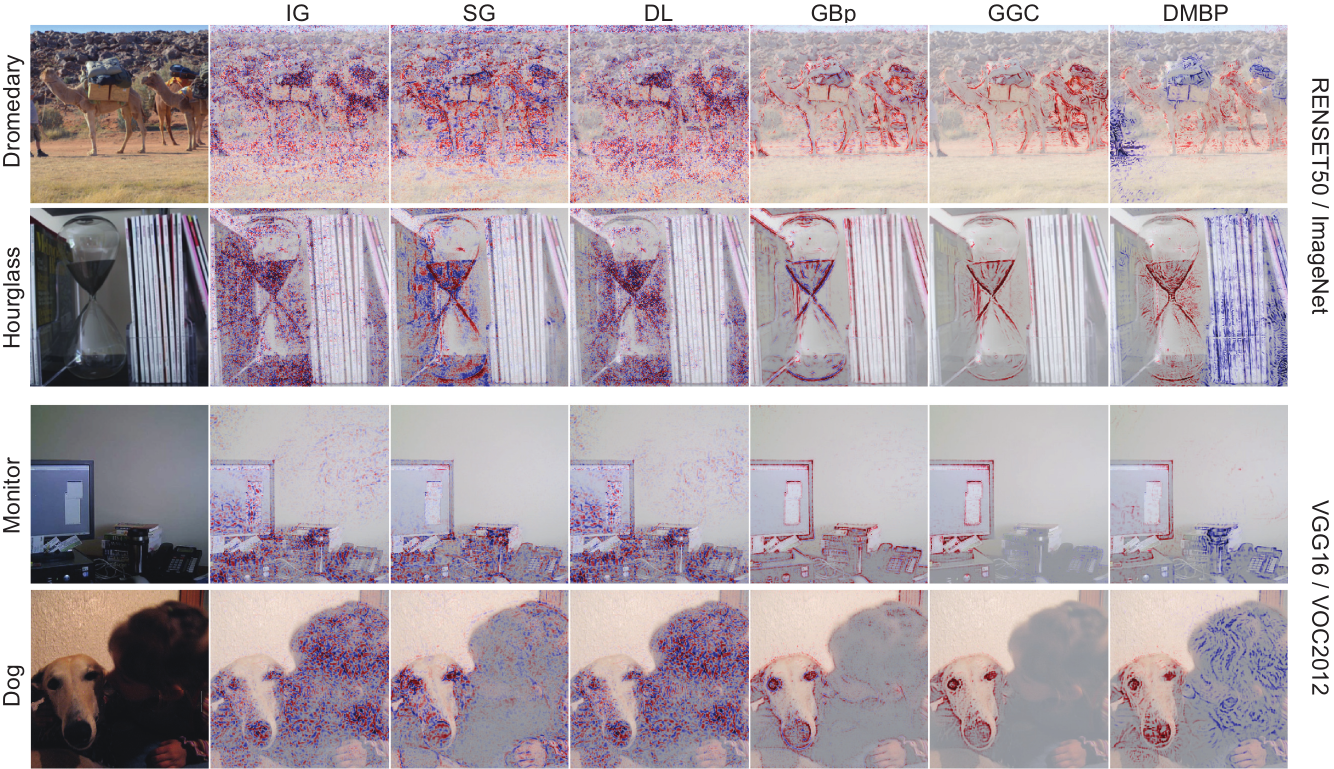}
\caption{{\bf Qualitative results for DMBP and alternative gradient-based approaches.}  More results provided in suppl. material.}
\label{fig:grad_based_com}
\end{figure*}

\tab{ablation} summarizes the results for the different evaluated metrics, datasets and models. As we can observe, ND yields significantly worse results compared to the different DMBP versions in most metrics. The reason is that positive, negative and nuisance factors are not disentangled in this case. This generates noisy results where attributions do not correctly identify the contribution of each pixel to the model's ouput. On the other hand, DMBP$^{+,-}$ consistently outperforms DMBP$^{+}$ in all the cases except for the IM metric in VGG16 over VOC. This improvement is because DMBP$^{+}$ ignores the negative factors during optimization. As a consequence, the discriminative patterns having a negative impact on the model's output are not correctly identified. This is clearly seen by observing the poor cIM values obtained by DMBP$^{+}$. Finally, DMBP$^{All}$  achieves comparable or better results than DMBP$^{+,-}$ in most cases. This demonstrates the importance of removing non-discriminative factors by minimizing the effect of the nuisance term ${y}^\sim$.

\subsection{Comparison with state of the art}
\label{sec:soa_comp}

\mypar{Comparison with gradient-based methods} As can be obzerved in \tab{soa_table}, DMBP consistently outperforms the rest of compared gradient-based approaches (Grad, IG, SG, BIG, DL, GBp and GGC). To provide further insights, Fig.~\ref{fig:grad_based_com} shows qualitative results obtained with DMBP and the alternative gradient-based methods with better performance. As shown, IG, SG and DL produce noisy visualization, in which positive and negative attributions are mixed. The reason is that these approaches do not explicitly disentangle the discriminative and non-relevant factors. In contrast, GBp uses hand-crafted backpropagation rules to identify image regions corresponding only to positive factors. 
However, this method produces attributions where pixels that do not belong to the target class are also assigned with positive attributions. Finally, GGC uses the coarse feature maps produced by GradCam to filter the attributions generated by GBp. Still, this strategy is not able to identify image pixels that have a negative contribution to the network output. In contrast to IG, SG and DL, our method produces more interpretable attribution maps by removing the non-discriminative factors. Additionally, compared to GGC and GBp, our approach correctly identifies factors that have a negative effect on the output.

\mypar{Comparison with other approaches} Results in \tab{soa_table} also demonstrate the advantages of DMBP over the approaches using intermediate layer information GradCam and FG. More concretely, DMBP obtains the best average performance and the top-scoring results in all the metrics except for VGG16 over ImageNet, where FG and GC outperform our approach. Nevertheless, these two methods generate attribution maps by upsampling the information extracted from intermediate layers. As shown in Fig.~\ref{fig:coarse_com}, this results in coarser and less informative visualizations than those obtained with our method. Whereas coarse attribution maps can be potentially applied to specific downstream tasks such as weakly-supervised object localization, reliable network intepretability requires fine-grained results providing detailed information about the visual cues that the model exploits during inference. As can be observed, this is the case for DMBP visualizations, which provide fine-grained pixel-level attributions identifying high-frequency information such as object edges or textures. To conclude, our method consistently outperforms the perturbation-based methods RISE and LPR. The latter also optimizes function derivatives for backpropagation. However, the modified gradients are used to generate binary masks corrupting the original image. In contrast, DMBP achieves better performance by using the modified gradients to explicitly disentangle positive, negative and nuisance factors.

\input{Tables/reinit_metric_table}

\mypar{Attributions sensitivity to layer reinitialization}
DMBP generates fine-grained attributions where visual cues such as edges can be clearly identified. The identification of this intuitive patterns is also observed in gradient-based approaches such as Gradient Backpropagation and Gradient GradCam. However, \cite{adebayo2018sanity} showed that these approaches suffer from a critical weakness: the attribution maps are not sensitive to the reinitialization of network parameters. As a consequence, they cannot be identifying the discriminative regions explaining the inference process. In order to evaluate if DMBP suffers from this limitation, we perform a sanity check proposed in \cite{adebayo2018sanity}. In particular, we compute the rank correlation between the original attribution maps and the ones generated by a network where the last layer parameters are randomly reinitialized using a normal distribution. \tab{reinit_metric} shows the obtained results. As can be observed, the high correlations obtained by GBp and GGC indicate that these methods generate attribution maps that are not sensitive to model reinitialization. In contrast, DMBP obtains almost a zero correlation for all cases, showing that the identified edges are truly dependent on network parameters.

%% file: Tables/ablation_table.tex
\begin{table}[t]
\centering
\resizebox{\linewidth}{!}{%
\begin{tabular}{cccccc}
\textbf{DB/Model} & \textbf{Metric} & \textbf{ND} & \textbf{DMBP$^+$} & \textbf{DMBP$^{+,-}$} & \textbf{DMBP$^{All}$} \\ \hline
\multirow{2}{*}{VOC/VGG16} & IM & .36 & \textbf{{[}.62{]}} & .51 & \textbf{.57} \\
 & cIM & .09 & .06 & \textbf{{[}.18{]}} & \textbf{.16} \\
\multirow{2}{*}{VOC/RN50} & IM & .30 & .58 & \textbf{.59} & \textbf{{[}.61{]}} \\
 & cIM & .15 & .15 & \textbf{{[}.22{]}} & \textbf{{[}.22{]}} \\
ImNet/VGG16 & IM & .24 & .37 & \textbf{.37} & \textbf{{[}.39{]}} \\
ImNet/RN50 & IM & .26 & .50 & \textbf{.51} & \textbf{{[}.57{]}}
\end{tabular}%
}
\caption{{\bf Results for DMBP variants optimized with different disentanglement losses.} See text for details. Best method is indicated with brackets. Second and best are indicated in bold.  }
\label{tab:ablation}
\end{table}

%% file: Tables/soa_table2.tex
\begin{table*}[t]
\centering
\resizebox{\textwidth}{!}{%
\begin{tabular}{cccccccccccccc}
\textbf{DB/Model} & \textbf{Metric} & \textbf{Grad} \cite{simonyan} & \textbf{IG} \cite{sundararajan2017} & \textbf{SG} \cite{smilkov2017} & \textbf{BIG} \cite{google2020} & \textbf{DL} \cite{shrikumar2017} & \textbf{GBp} \cite{springenberg2015} & \textbf{GGC} \cite{selvaraju2019} & \textbf{LPR} \cite{yang2019} &\textbf{RISE} \cite{petsiuk2018} & \textbf{FG} \cite{srinivas2019} & \textbf{GC} \cite{selvaraju2019} & \textbf{DMBP} \\ \hline
\multirow{2}{*}{VOC/VGG} & IM & 0.36 & 0.43 & \textbf{0.55} & 0.36 & 0.41 & 0.32 & 0.35 & 0.40 & 0.41 & 0.28 & 0.35 & \textbf{{[}0.57{]}} \\
 & cIM & 0.09 & 0.06 & 0.06 & 0.08 & 0.07 & 0.06 & 0.12 & 0.07 & \textbf{0.13} & 0.10 & \textbf{0.13} & \textbf{{[}0.16{]}} \\
\multirow{2}{*}{VOC/RN50} & IM & 0.30 & 0.36 & 0.47 & 0.27 & 0.34 & 0.39 & 0.50 & 0.40 & \textbf{0.51} & 0.44 & \textbf{0.51} & \textbf{{[}0.61{]}} \\
 & cIM & 0.15 & 0.16 & 0.15 & 0.15 & 0.15 & 0.13 & 0.19 & 0.14 & 0.21 & 0.14 & \textbf{0.21} & \textbf{{[}0.22{]}} \\
INet/VGG & IM & 0.23 & 0.28 & 0.38 & 0.21 & 0.30 & 0.32 & 0.36 & 0.28 & 0.28 & \textbf{0.43} & \textbf{{[}0.48{]}} & 0.41 \\
INet/RN50 & IM & 0.26 & 0.30 & 0.40 & 0.27 & 0.29 & 0.43 & 0.50 & 0.33 & 0.33 & 0.53 & \textbf{0.55} & \textbf{{[}0.56{]}} \\ \hline
\textbf{} & Avg & 0.23 & 0.27 & 0.34 & 0.22 & 0.26 & 0.28 & 0.34 & 0.27 & 0.31 & 0.32 & \textbf{0.37} & \textbf{{[}0.42{]}}
\end{tabular}%
}
\caption{{\bf Results obtained by DMBP and other state-of-the-art methods.} Metrics are shown for all the evaluated datasets and network models. Best method is indicated with brackets. Second and best are indicated in bold.}
\label{tab:soa_table}
\end{table*}


%% file: Tables/reinit_metric_table.tex
\begin{table}[t]
\centering
\resizebox{0.85\linewidth}{!}{%
\begin{tabular}{ccccc}
\textbf{} & \multicolumn{2}{c}{\textbf{VOC}} & \multicolumn{2}{c}{\textbf{IMNet}} \\ \cline{2-5} 
\textbf{} & \textbf{VGG16} & \textbf{RN50} & \textbf{VGG16} & \textbf{RN50} \\ \hline
\textbf{GBp \cite{springenberg2015}} & 0.99 & 0.99 & 0.99 & 0.99\\
\textbf{GGC \cite{selvaraju2019}} & 0.25 & 0.28 & 0.35 & 0.45 \\
\textbf{DMBP (Ours)} & \textbf{0.02} & \textbf{-0.03} & \textbf{0.00} & \textbf{0.00}
\end{tabular}%
}
\caption{Correlation between the attributions obtained from the original network and for one with reinitialized parameters. }
\label{tab:reinit_metric}
\end{table}

%% file: supplementary.tex
\section{Incorporating Bias Terms} 
\label{sec:sup_bias_details}
In the following, we provide a simple example to show that the output for a ReLU neural network with bias terms can be computed as in \Eq{bias_terms}. In particular, let us consider the output for a network with $3$ hidden layers:
\begin{equation}
    \label{eq:example_mlp}
    y = \mathbf{w}^T \phi \bigg(\mathbf{W}_{3} \bigg[ \phi(\mathbf{W}_{2} \big[ \phi(\mathbf{W}_{1} \mathbf{x} + \mathbf{b}_1) \big] + \mathbf{b}_2)\bigg] \bigg)
\end{equation}
For the sake of simplicity and without loss of generality, we assume that no bias is applied to the last hidden layer. Given the defined network, the output $y$ can be obtained by the sequential computation of the intermediate layers as:

\begin{align}
\mathbf{h}_1 &= \mathbf{W}_{1} \mathbf{x} \nonumber \\ 
\mathbf{h}_2 &= \mathbf{W}_{2} \phi(\mathbf{h}_1+\mathbf{b}_1) = \mathbf{\hat{W}}_{2} \mathbf{W}_{1} \mathbf{x} + \mathbf{\hat{W}}_{2} \mathbf{b}_1 \nonumber \\ 
\mathbf{h}_3 &= \mathbf{W}_{3}  \phi(\mathbf{h}_2+\mathbf{b}_2) = \mathbf{\hat{W}}_{3} \mathbf{\hat{W}}_{2} \mathbf{W}_{1} \mathbf{x} +  \mathbf{\hat{W}}_{3} \mathbf{\hat{W}}_{2} \mathbf{b}_1 + \mathbf{\hat{W}}_{3} \mathbf{b}_2  \nonumber \\ 
y &= \mathbf{w}^T \phi(\mathbf{h}_3) = \mathbf{\hat{w}}^T \mathbf{\hat{W}}_{3} \mathbf{\hat{W}}_{2} \mathbf{W}_{1} \mathbf{x} + \mathbf{\hat{w}} \mathbf{\hat{W}}_{3} \mathbf{\hat{W}}_{2} \mathbf{b}_1 + \mathbf{\hat{w}}^T \mathbf{\hat{W}}_{3} \mathbf{b}_2 
\end{align}
where $\mathbf{\hat{W}}_{l} = \mathbf{W}_{l} \diag(\mathcal{H}(\mathbf{h}_{l-1}  + \mathbf{b}_{l-1}))$, and $\mathbf{\hat{w}} = \mathbf{\hat{w}}  \diag(\mathcal{H}(\mathbf{h}_{3}))$. By using the same procedure for any network with an arbitrary depth, it is clear that the network output can be expressed by using \Eq{bias_terms} in the main paper:
\begin{align}
    y &=  \mathbf{\hat{w}}^T \bigg[  \mathbf{\hat{W}}_{L} \dots \mathbf{W}_{1} \bigg] \mathbf{x} + \sum_{l=2}^L \mathbf{\hat{w}}^T \bigg[  \mathbf{\hat{W}}_{L} \dots \mathbf{\hat{W}}_{l} \bigg] \mathbf{b}_{l-1},
\end{align}

\section{DMBP Optimization}
In Algorithm \ref{alg:pseudocode}, we provide the pseudocode for computing attribution maps with DMBP over CNNs.
\label{sec:sup_pseudocode}

\input{Others/DMBP_algorithm}

\clearpage

\subsection{Initializing filter masks $\mathbf{\Sigma}_l$}
\label{sec:sup_initialization}
In our preliminary experiments, we observed that a random initialization of parameters $\mathbf{\Sigma}_l$ produces suboptimal results in some cases. In order to provide a better initial point for DMBP optimization, we use the following procedure.

Starting from the last layer $L$, we perform two parallel backward passes computing the  positive and negative gradients $\nabla_{\mathbf{x}} F^+(\mathbf{x})$ and $\nabla_{\mathbf{x}} F^-(\mathbf{x})$ using $\mathbf{\Sigma}_l$ and $\mathbf{I}-\mathbf{\Sigma}_l$, respectively. During the backpropagation process, each element $i$ of the filter masks is initialized for each layer $l$ as:

\[ \mathbf{\Sigma}^i_l = \begin{cases} 
          \text{sigmoid}(2) \hspace{5mm} \text{iff} \hspace{4mm} \nabla_{\mathbf{h}_l}^i F^+(\mathbf{x}) > 0 \hspace{2mm}\text{and}\hspace{2mm} \nabla_{\mathbf{h}_l}^i F^-(\mathbf{x}) > 0\\
          \text{sigmoid}(-2) \hspace{2mm} \text{iff} \hspace{4mm} \nabla_{\mathbf{h}_l}^i F^+(\mathbf{x}) < 0 \hspace{2mm}\text{and}\hspace{2mm} \nabla_{\mathbf{h}_l}^i F^-(\mathbf{x}) < 0 \\
          0.5 \hspace{17mm} \text{otherwise}
       \end{cases}
\]
where $\nabla_{\mathbf{h}_l} F^+(\mathbf{x})$ is the positive gradient of the output \wrt the activations $\mathbf{h}_l$. Similarly, $\nabla_{\mathbf{h}_l} F^-(\mathbf{x})$ is the negative gradient obtained by using $\mathbf{I}-\mathbf{\Sigma}_l$ during backpropagation.

The proposed initialization procedure is motivated by the fact that, ignoring the bias terms, the positive and negative outputs $y^+(\boldsymbol \sigma)$ and $y^-(\boldsymbol \sigma)$ can be expressed as:
\begin{equation}
y^+(\boldsymbol \sigma)= [\nabla_{\mathbf{h}_l} F^+(\mathbf{x}) \odot \mathbf{\Sigma}_l]^T \mathbf{h}_l \hspace{2.5mm} , \hspace{2.5mm} y^-(\boldsymbol \sigma)=[\nabla_{\mathbf{h}_l} F^-(\mathbf{x}) \odot (\mathbf{I}-\mathbf{\Sigma}_l)]^T \mathbf{h}_l,
\end{equation}
where $\mathbf{h}_l \geq 0$ is the result of a ReLU operation and thus, it is always positive. As a consequence, we can locally increase the value of $y^+(\boldsymbol \sigma)$ by assigning low values to elements $\mathbf{\Sigma}_l^i$ when  $\nabla_{\mathbf{h}_l}^i F^+(\mathbf{x})$ is negative. Additionally, large values for $\mathbf{\Sigma}_l^i$ must be assigned to elements where $\nabla_{\mathbf{h}_l}^i F^+(\mathbf{x})$ is positive. Given that the negative score $y^-(\boldsymbol \sigma)$ is computed using $\mathbf{I}-\mathbf{\Sigma}_l$ during backpropagation, our initialization procedure also takes into account the negative gradient $\nabla_{\mathbf{h}_l}^i F^-(\mathbf{x})$. In particular, we decrease $y^-(\boldsymbol \sigma)$ by assigning a high value to $1-\mathbf{\Sigma}^i_l$ when $\nabla_{\mathbf{h}_l}^i F^-(\mathbf{x})$ is negative. Similarly, we set a low value to $1-\mathbf{\Sigma}^i_l$ for positive values of $\nabla_{\mathbf{h}_l}^i F^-(\mathbf{x})$.
In cases where positive and negative values are not consistent (\ie $\nabla_{\mathbf{h}_l}^i F^+(\mathbf{x})$ and $\nabla_{\mathbf{h}_l}^i F^-(\mathbf{x})$ do not have the same sign), we simply set $\mathbf{\Sigma}^i_l$ to $0.5$.

\section{Additional Qualitative Results} 
\label{sec:sup_qualres}
In the following, we show additional qualitative results comparing DMBP with previous gradient-based approaches. For each example, we show the attribution maps (third row) and the two linear mappings producing the positive and negative pixel attributions when they are multiplied by the image (first and second rows). For DMBP, the latter are explicitly computed during optimization. For the rest of methods, they are computed by decomposing the obtained linear mapping according to the sign of the obtained pixel-level attributions. The visualization of these linear mappings allow to understand the low-level information (\ie colors, edges, textures) that produces the attributions for each pixel. From the reported results, we can extract the same conclusions than the ones discussed in \sect{soa_comp}.

\input{supplementary_figures}

\clearpage

%% file: Others/DMBP_algorithm.tex
\begin{algorithm}[h]
\small
\caption{Attribution Map Compuptation with DMBP}
\label{alg:dmbp_pseudocode}
\begin{algorithmic}[1]
 \renewcommand{\algorithmicrequire}{\textbf{Input:}}
 \renewcommand{\algorithmicensure}{\textbf{Output:}}
 \REQUIRE Image: $\mathbf{x}$, Network: $F$, Target label: $y$
 \ENSURE  Attribution Map: $\mathbf{a}$
 \STATEx \hspace{-4mm} \textit{// Initialization}
 \STATE Initialize filter masks $\mathbf{\Sigma}_l$ for all $l$ as described in \app{sup_initialization}
 \STATEx \hspace{-4mm} \textit{// Initial Forward Pass}
 \STATE Compute $y=F(\mathbf{x})$
 \STATEx \hspace{-4mm} \textit{// Linearize the CNN function with \Eq{bias_terms} and minimize \Eq{loss} with gradient descent}
 \FOR {$i = 1$ to \textit{iters}}
 
  \STATEx \hspace{3mm} \textit{// Compute positive term} $y(\boldsymbol \sigma)^+$ \textit{using \Eq{pos_term} //}
  \STATE  $\nabla^+_\mathbf{x}F(\mathbf{x}) \leftarrow$ Backward pass with $\mathbf{\Sigma}_l \odot \nabla_{\mathbf{h}_l} F(\mathbf{x})$
  \STATE $y(\mathbf{x})^+=\nabla^+_\mathbf{x}F(\mathbf{x})^T \mathbf{x} + \sum_{l=0}^L \nabla^+_\mathbf{b_l} F(\mathbf{x})^T \mathbf{b_l}$
  \STATEx
  \STATEx \hspace{3mm} \textit{// Compute negative term} $y(\boldsymbol \sigma)^-$ \textit{using \Eq{pos_term} //}
  \STATE  $\nabla^-_\mathbf{x}F(\mathbf{x}) \leftarrow$ Backward pass with $(1-\mathbf{\Sigma}_l) \odot \nabla_{\mathbf{h}_l} F(\mathbf{x})$
  \STATE $y(\mathbf{x})^-=\nabla^-_\mathbf{x}F(\mathbf{x})^T \mathbf{x} + \sum_{l=0}^L \nabla^-_\mathbf{b_l} F(\mathbf{x})^T \mathbf{b_l}$
  \STATEx
  \STATEx \hspace{3mm} \textit{// Compute nuisance term} $y(\mathbf{x})^\sim$ \textit{//}
  \STATE $y^\sim(\boldsymbol \sigma) = y -  y^+(\boldsymbol \sigma) - y^-(\boldsymbol \sigma)$
  \STATEx
  \STATEx \hspace{3mm} \textit{// Loss optimization} \textit{//}
  \STATE Compute $\mathcal{L}=y^-(\boldsymbol \sigma) - y^+(\boldsymbol \sigma) + \lVert y^\sim(\boldsymbol \sigma)\rVert_1$
  \STATE Compute loss gradients $\nabla_{\mathbf{\Sigma}_{l}} \mathcal{L}$ for all $\mathbf{\Sigma}_l$
  \STATE Update $\mathbf{\Sigma}_{l}$ for all $l$
\ENDFOR
\STATEx \hspace{-6mm} \textit{// Return attribution map} $\mathbf{a}$ \textit{//}
\STATE $\mathbf{a} = \nabla^+_\mathbf{x}F(\mathbf{x}) \odot  \mathbf{x} + \nabla^-_\mathbf{x}F(\mathbf{x}) \odot  \mathbf{x}$
\end{algorithmic} 
\label{alg:pseudocode}
\end{algorithm}

%% file: supplementary_figures.tex
\begin{figure*}[ht]
\label{fig:vgg_voc}
\centering
\includegraphics[width=0.95\linewidth]{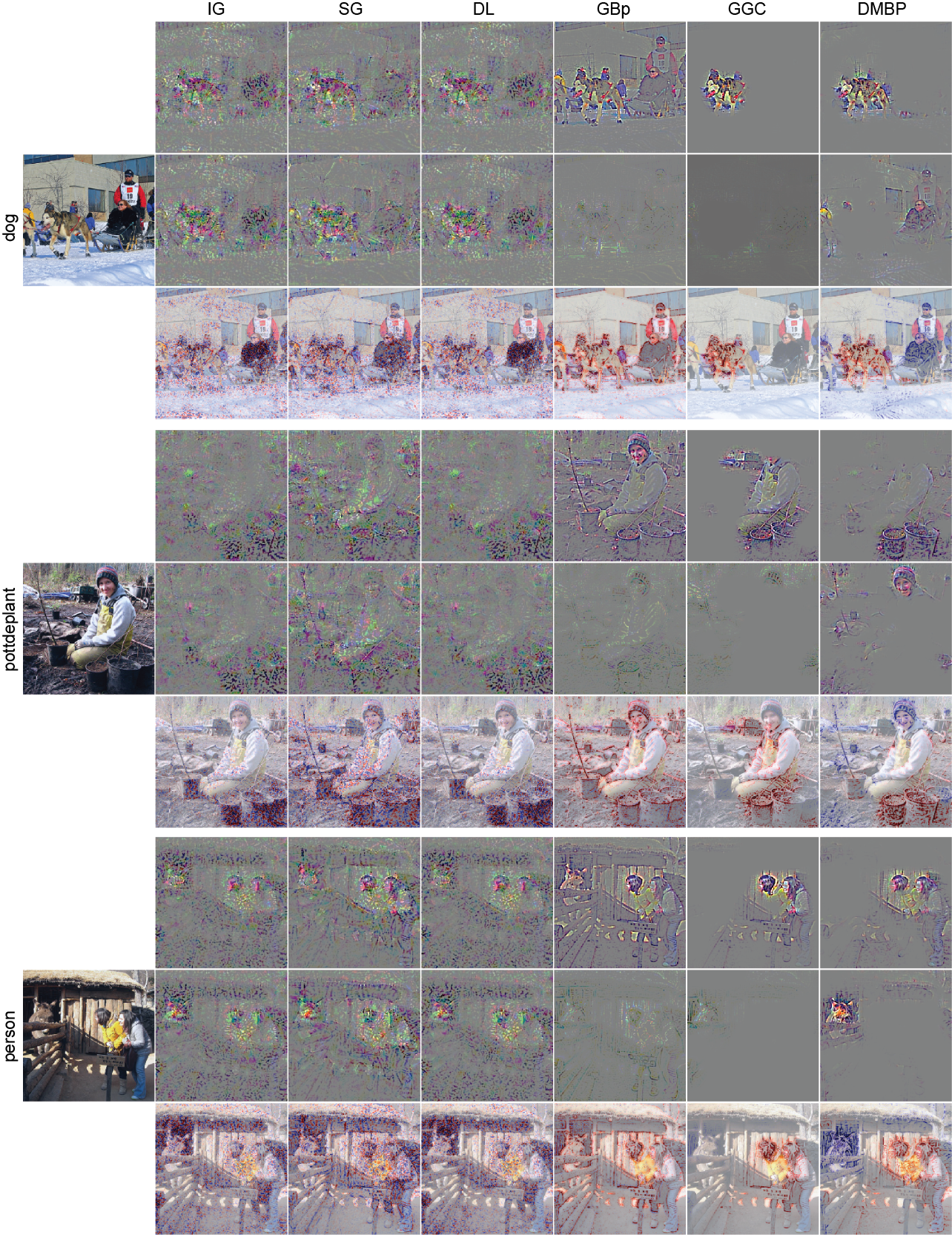}
\caption{{\bf Qualitative results for DMBP and alternative gradient-based approaches}. VGG16 applied over VOC images.
}
\end{figure*}

\begin{figure*}[ht]
\label{fig:vgg_voc2}
\centering
\includegraphics[width=0.95\linewidth]{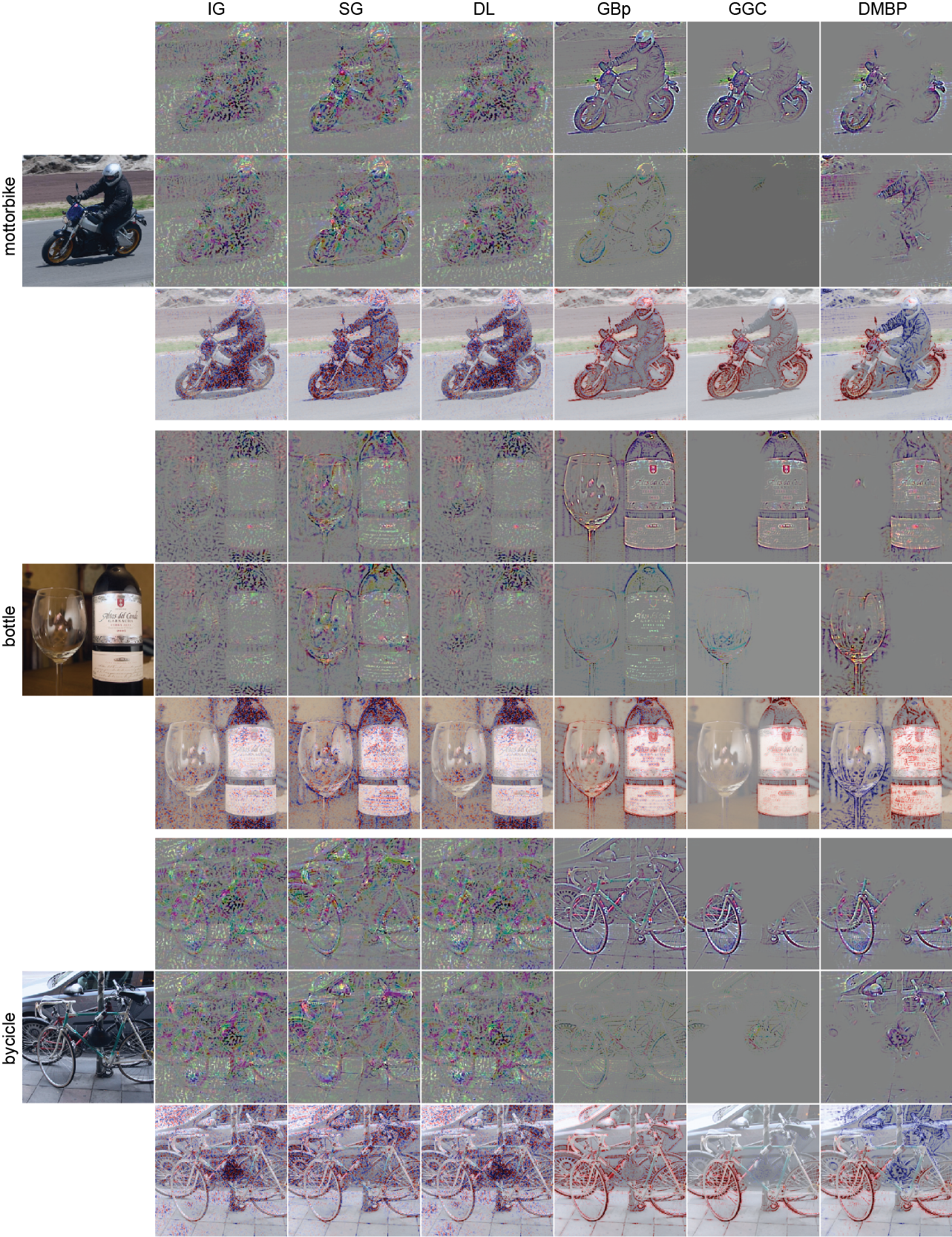}
\caption{{\bf Qualitative results for DMBP and alternative gradient-based approaches}. VGG16 applied over VOC images.
}
\end{figure*}

\begin{figure*}[ht]
\label{fig:vgg_resnet}
\centering
\includegraphics[width=0.95\linewidth]{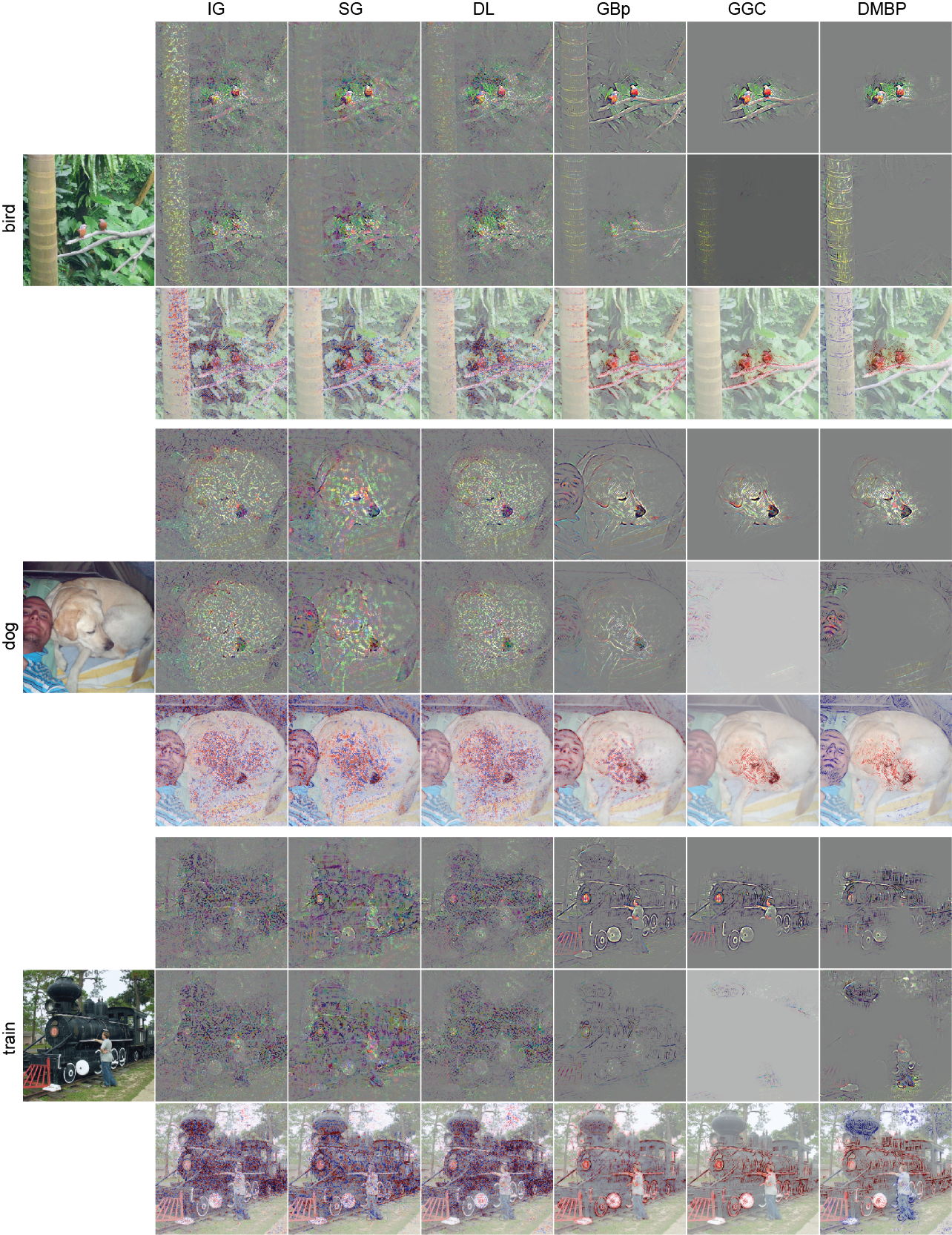}
\caption{{\bf Qualitative results for DMBP and alternative gradient-based approaches}. ResNet50 applied over VOC images.
}
\end{figure*}

\begin{figure*}[ht]
\label{fig:vgg_resnet2}
\centering
\includegraphics[width=0.95\linewidth]{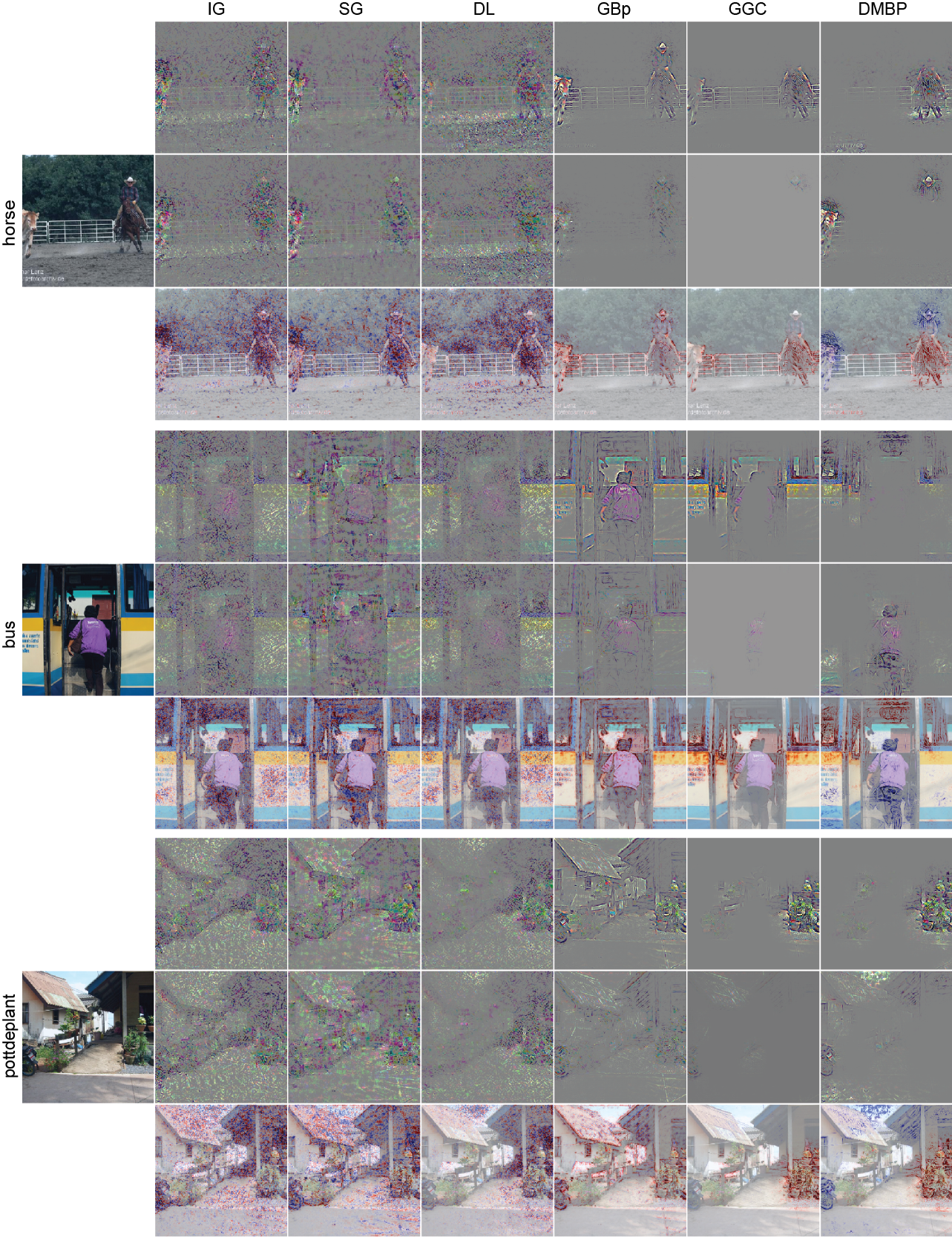}
\caption{{\bf Qualitative results for DMBP and alternative gradient-based approaches}. ResNet50 applied over VOC images.
}
\end{figure*}

\begin{figure*}[ht]
\label{fig:resnet50_imnet}
\centering
\includegraphics[width=0.95\linewidth]{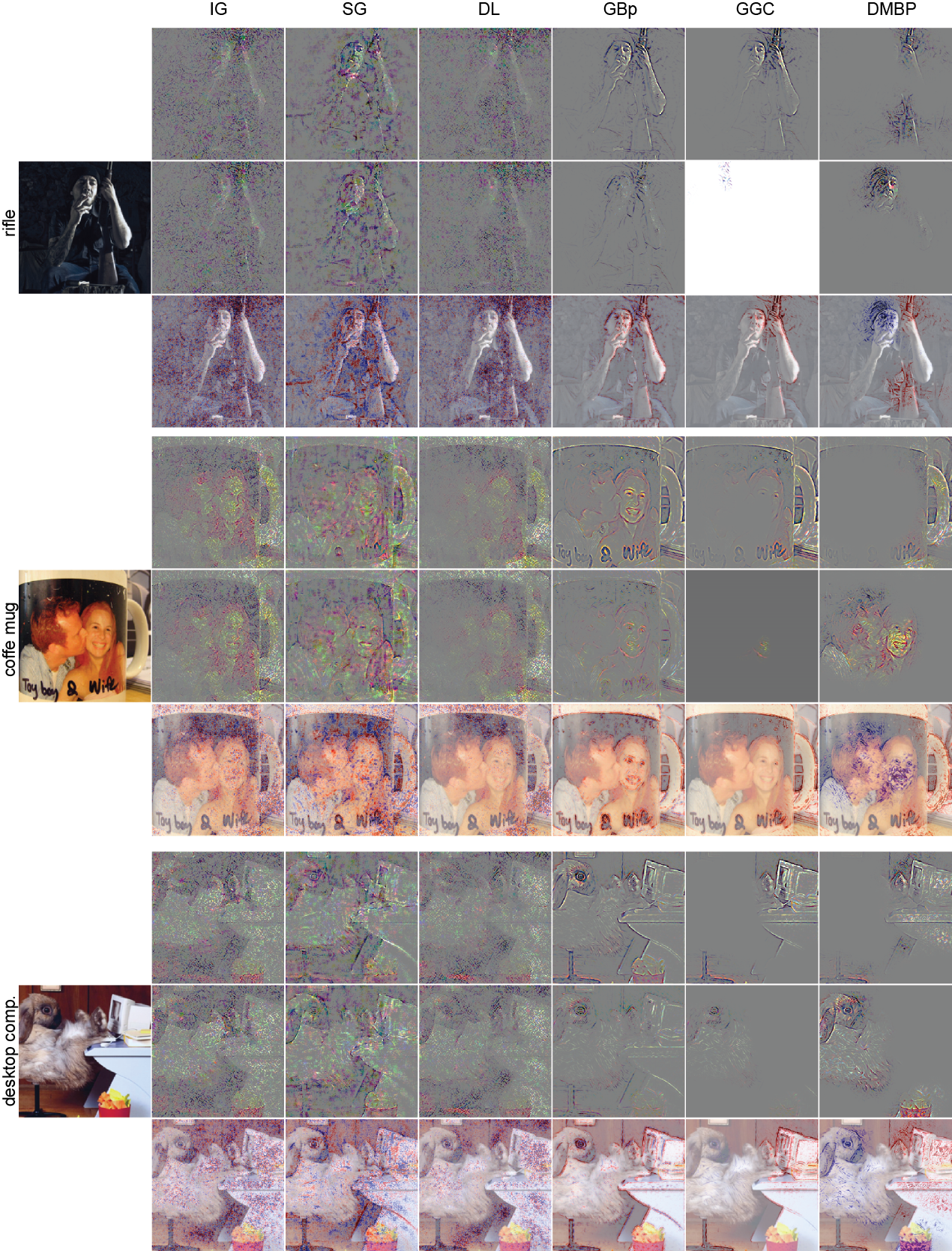}
\caption{{\bf Qualitative results for DMBP and alternative gradient-based approaches}. ResNet50 applied over ImageNet images.
}
\end{figure*}

\begin{figure*}[ht]
\label{fig:resnet50_imnet2}
\centering
\includegraphics[width=0.95\linewidth]{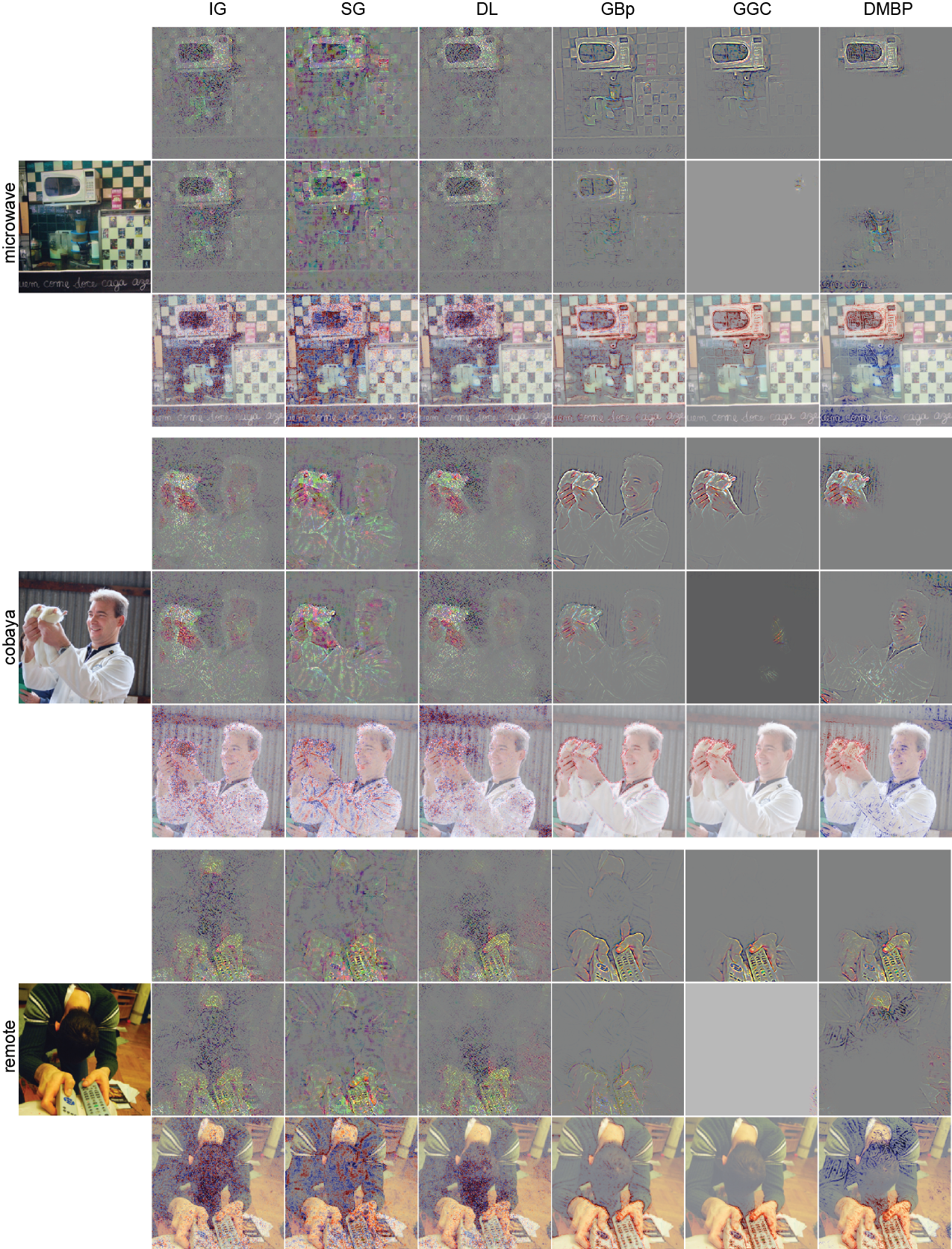}
\caption{{\bf Qualitative results for DMBP and alternative gradient-based approaches}. ResNet50 applied over ImageNet images.
}
\end{figure*}

\begin{figure*}[ht]
\label{fig:vgg16_imnet}
\centering
\includegraphics[width=0.95\linewidth]{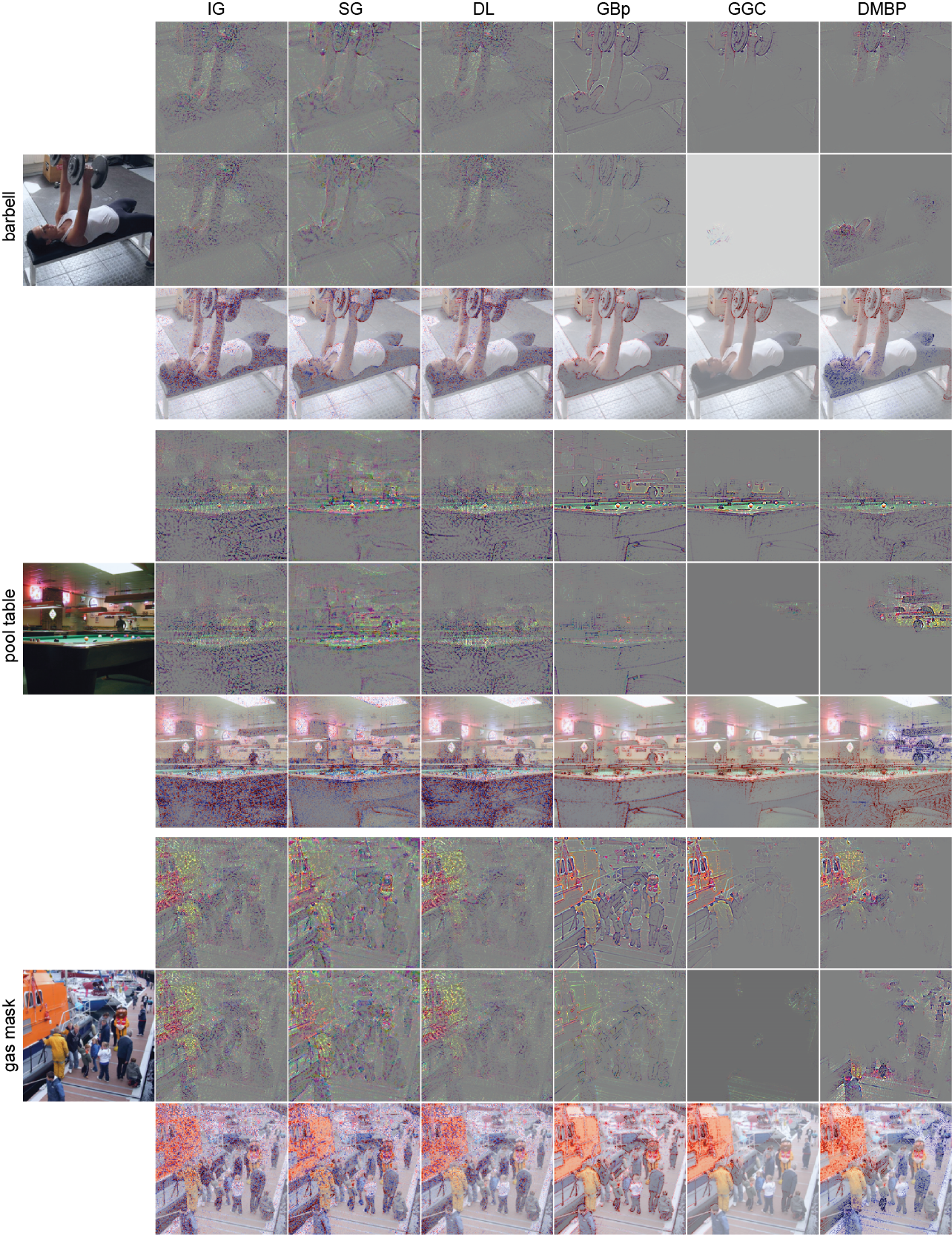}
\caption{{\bf Qualitative results for DMBP and alternative gradient-based approaches}. VGG16 applied over ImageNet images.
}
\end{figure*}

\begin{figure*}[ht]
\label{fig:vgg16_imnet2}
\centering
\includegraphics[width=0.95\linewidth]{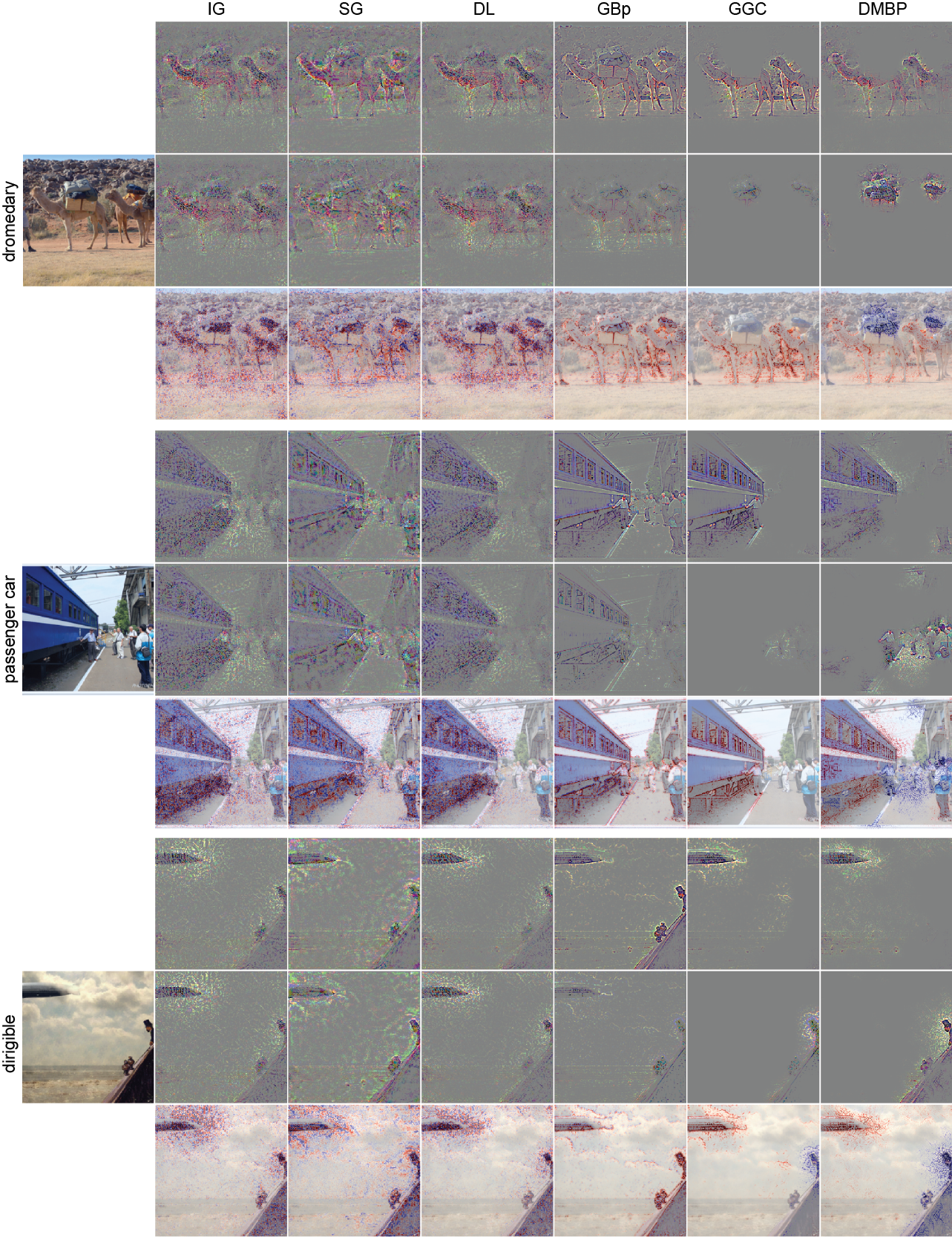}
\caption{{\bf Qualitative results for DMBP and alternative gradient-based approaches}. VGG16 applied over ImageNet images.
}
\end{figure*}